\documentclass{bmvc2k}
\usepackage{floatrow}
\newfloatcommand{capbtabbox}{table}[][\FBwidth]

\usepackage{blindtext}
\usepackage{array}
\usepackage{booktabs}
\DeclareMathOperator*{\argmax}{arg\,max}
\title{\textit{From Benedict Cumberbatch to Sherlock Holmes:} Character Identification in TV series without a Script}

\addauthor{Arsha Nagrani}{arsha@robots.ox.ac.uk/}{1}
\addauthor{Andrew Zisserman}{az@robots.ox.ac.uk/}{1}

\addinstitution{Visual Geometry Group, \\
Department of Engineering Science,
University of Oxford, UK }

\runninghead{Nagrani, Zisserman}{From Benedict Cumberbatch to Sherlock Holmes}

\begin{document}

\maketitle

\begin{abstract}

The goal of this paper is the automatic identification of characters in TV and feature film material. 
In contrast to standard approaches to
this task, which rely on the weak supervision afforded by transcripts and subtitles, we propose a
new method requiring only a cast list. This list is used to obtain
images of actors from freely available sources on the web, providing a form of partial supervision for this task. In using images of {\em actors} to recognize {\em
characters}, we make the following three contributions: (i) We
demonstrate that an automated semi-supervised learning approach is
able to adapt from the actor's face to the character's face, including
the face {\em context} of the hair; (ii) By building voice models for
every character, we provide a bridge between frontal faces (for which
there is plenty of actor-level supervision) and profile (for which
there is very little or none); and (iii) by combining face context and
speaker identification, we are able to identify characters with
partially occluded faces and extreme facial poses.

Results are presented on the TV series `Sherlock' and the feature film
`Casablanca'. We achieve the state-of-the-art on the Casablanca
benchmark, surpassing previous methods that have used the stronger
supervision available from transcripts.

\end{abstract}

\section{Introduction}
\label{sec:intro}
\vspace{-5pt}
One of the long-term objectives in computer vision is video understanding, for which 
an important step is the ability to recognize peoples' identities under unconstrained conditions. 
In this area, 
TV dramas, sit-coms and feature films have provided a strong test bed,
where the goal has been to recognize characters
in video using their faces (albeit usually only frontal faces). 
Achieving this goal is crucial to story understanding, and perhaps more directly, to generate metadata for indexing and
intelligent fast forwards (e.g.\ ``fast forward to scenes when Sherlock speaks to John'').
Since the work of Everingham {\it et al.}~\cite{Everingham06a}, most
methods have made use of a transcript aligned with the subtitles to provide
weak supervision for the task~\cite{Bojanowski13,Cinbis11,Cour09,Cour10,Cour11,Kostinger11,Parkhi15a,Ramanathan14,Sivic09,Tapaswi12,Wohlhart11}. 

\begin{figure*}[ht] 
\centering 
\includegraphics[width=1\textwidth]{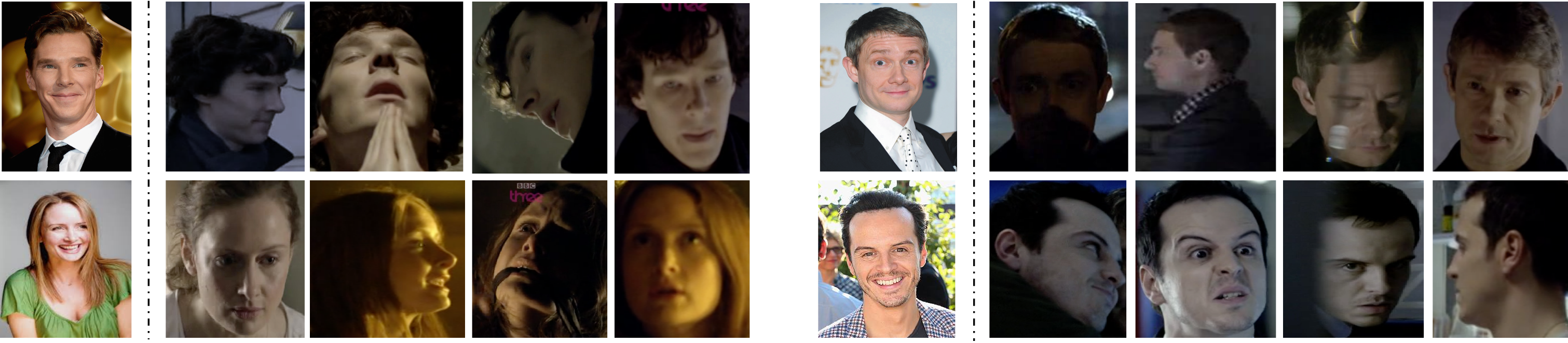} 
\small{\caption{Examples of occurrences of actors in source (leftmost image in each set) and target domains for four identities. The source images (from web images) can differ from the target images (from the TV material) in hairstyle, makeup, lighting, and viewpoint.}}
\label{fig:domains} 
\vspace{-5pt}
\end{figure*}

In this paper, our goal is also to recognize characters from video material; the
novelty of our approach is that we eschew the use of transcripts and
subtitles in favor of a simpler solution that requires only the cast list. The key idea we
explore, is that it is possible to get {\em partial} supervision from
images of the actor's faces that are easily available using web image search
engines such as Google or Bing's Image Search. This supervision is only
partial, however, because of the disparity between appearances of actors and their corresponding
characters. Images of actors available online are typically from red-carpet
photos or interviews, whereas a character can have a substantially
different hair style, make-up, viewpoint (extreme profile, rather
than the near frontal of actor interviews), and possibly even some facial
prosthetics. In essence, the problem is one of domain adaptation, as shown
in Figure~\ref{fig:domains}.

Using only actor-level supervision to recognize characters, we
make the following contributions:
(i) We demonstrate that an automated semi-supervised learning approach is
able to adapt from the actor's face to the character's face, including
the face {\em context} of the hair;
(ii) We use active speaker detection and speaker identification to build voice models for every character, and show that such models can act as a bridge between frontal faces (for which there is
plenty of actor-level supervision) and profile (for which there is very little
or none); and (iii) by combining face context and speaker identification,
we are able to identify characters with
partially occluded faces and further profile faces.
The outcome is a set of characters labelled in both frontal and
profile viewpoints.

We demonstrate the method on two datasets: the TV series `Sherlock',
episodes 1--3, which will be used as the running example in this
paper; and the feature film `Casablanca' which is used as a benchmark
for this task. On Casablanca, we achieve state-of-the-art results,
surpassing previous methods that have used the stronger supervision
available from transcripts and subtitles.

We pursue a transcript-free approach for a number of
reasons. Transcripts, typically obtained from IMDB or fan sites, are
difficult to find for all films or entire seasons of a TV
series. They also tend to come in a variety of different styles or
formats, ensuring much work must go into standardizing
formats. 
In contrast, cast lists are easy to obtain for all episodes of a show
from IMDB. Cast lists can also be obtained visually from the credits
(using OCR) and hence no external annotation is required. We can also
aim to label every credited character, and not just principal
characters or those with speaking parts (as is typically done in approaches
using transcripts).
\vspace{-5pt}

\vspace*{-0.5cm}
\subsection{Related Work}
\vspace{-5pt}
When weak supervision is available from transcripts, then the
problem can be cast as one of ambiguous labelling~\cite{Cour09} or
Multiple Instance Learning (MIL), as employed
by~\cite{Bojanowski13,Kostinger11,Wohlhart11,Yang05}. 

In our case, where the partial supervision is provided by web images
given the actor's name, the approach is related to the on-the-fly
retrieval of~\cite{Parkhi12b}, although previous work has not
considered the domain adaptation aspect developed here.  In order to
fully solve the domain adaptation problem, we leverage the use of the
audio track to classify speech segments extracted automatically from
the video. While speaker detection is used to obtain labels from
transcripts and subtitles, to the best of our knowledge very few 
works~\cite{Tapaswi12,rouvier2014scene} have attempted to build
voice models to aid automatic labelling of tracks. By propagating labels from one modality to another (facetracks to audio segments and vice versa), we use a co-training approach similar to that outlined in~\cite{blum1998combining}.

Most previous methods for automated
character labelling have concentrated on frontal faces, with
only~\cite{Bojanowski13,Parkhi15a,Ramanathan14,Sivic09,Tapaswi12}
going beyond frontal faces to also consider
profiles.~\cite{zhang2015beyond} used poselet-level person recognizers in order to deal with large variations in pose, while~\cite{Ramanan07b} explicitly employed the hair as a measure
for tracking and was thus able to extend some frontal tracks to
profiles, though did not detect and track purely profile faces. The
consequence of this is that the non-profile methods ignore (and thus do not
attempt to identify) characters that appear in profile views. Those
that hav{}e attempted to also identify profile
views~\cite{Bojanowski13,Parkhi15a,Ramanathan14,Sivic09,Tapaswi12},
have often found inferior performance in profile, compared to frontal,
due to an inability to learn profile face classifiers when primarily
frontal faces are available for supervision. We show that voice models can be used to overcome this problem. 

\vspace{-5pt}
\begin{figure*}[ht] 
\centering 
\includegraphics[width=1\textwidth]{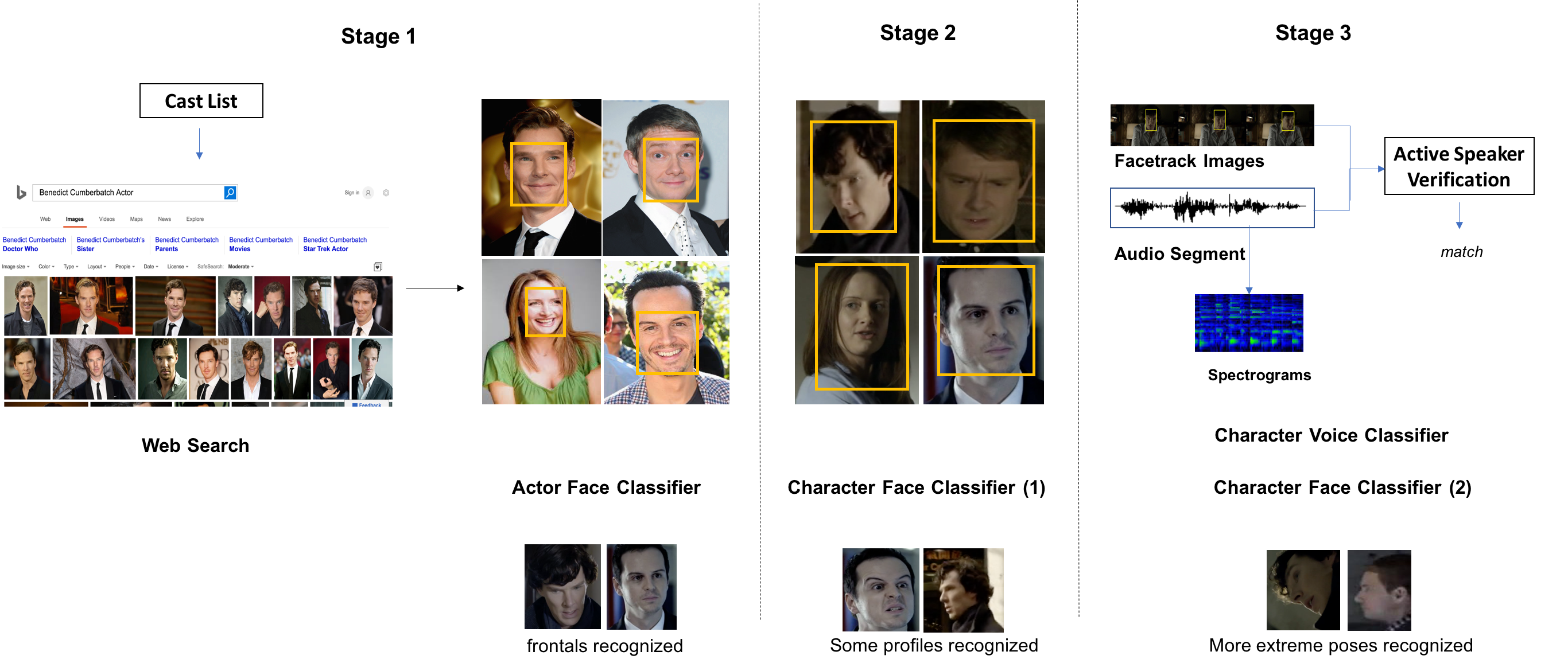} 
\caption{\small{Overview of the three stages of the approach.}}
\label{fig:overview} 
\vspace{-8pt}
\end{figure*}
  
\vspace*{-0.5cm}
\section{Overview}
\vspace{-5pt}
As is standard in almost all previous approaches for this
task~\cite{Bojanowski13,Cinbis11,Cour09,Cour10,Cour11,Kostinger11,Parkhi15a,Sivic09,Tapaswi12,Wohlhart11},
the basic unit of labelling for our task is a face track extracted from the
video. Starting from images of the actors available online, we build a
series of multiclass SVM classifiers, each relying on the output of
the previous classifier for supervision. Using this semi-supervised
technique, we can achieve robustness to pose, lighting, expression and
partial occlusions with very little initial supervision.

Our goal is to label all the face tracks (both of frontal and profile faces) for the actors/characters given in the cast list. The approach proceeds in
three main stages (shown in Figure~\ref{fig:overview}): \\
\textbf{1.\ Actor face classifier:} 
The first stage is to obtain initial labels for the face tracks
from actor images alone. Starting from the cast list, images of the
actor's faces are collected from search engine results available on
the web (Section \ref{Supervision}). A facial classifier is then
trained for each actor from these images, and used to classify the
face tracks. As demonstrated in Figure~\ref{fig:domains}, 
actor images have fundamental differences in pose,
lighting, expression and make up to the target images, and consequently the
face tracks labelled correctly at this stage are mostly limited to 
the relatively easy frontal images.

\noindent{\textbf{2.\ Character face classifier:}} 
The second stage involves building face classifiers for each
\textit{character} (Section~\ref{context}). Since some initial labels are available from 
stage~1 (from confident actor face classifiers), we can now train this character classifier on images from the video
directly. This allows us to learn the face context
(including the hair) of the character. The face tracks that the actor face classifier is not confident for, are then
reclassified with these character face classifiers. The outcome is that
some profile and partially occluded faces are now recognized correctly.

\noindent{\textbf{3.\ Character voice and face classifiers:}} 
The final stage involves adding the speech modality to improve results
further (Section \ref{Voice}). First, face tracks that are speaking are
determined using audio-visual correlation. Second, voice classifiers are
built for each character using the face labels provided by stage 2
(once again, labels are propagated from~\textit{confident} character face classifiers). The face tracks for which
the character face classifier is not confident are then 
reclassified by their voice. This
corrects the labels of some of the profile face tracks corresponding to
speaking characters. A final character 
face classifier is then trained including these
newly corrected labels, and all tracks reclassified by this
voice augmented character face classifier.
In practice, profile faces not detected
in stage~2, as well as those corresponding to speaking characters with
very little actor level supervision, are identified in this stage.


\vspace{-5pt}
\section{Face Classifiers} 
\vspace{-5pt}
We learn three different face classifiers. 
The first (actor
face classifier) is trained on images
obtained from the web and generates an initial classification
of the face tracks.
These classified face tracks are then used 
in a second round of training to learn
the character's appearance (character face classifier 1). Finally, additional
supervision is obtained from the speaker classifier, and the character
classifier retrained (character face classifier 2).
\vspace{-5pt}

\subsection{Actor Face Supervision} \label{Supervision}
\vspace{-5pt}
A dataset of actor images is obtained automatically using the following steps: \\
\noindent{\textbf{1. Cast lists:}}
 Cast lists are obtained for every episode from the Internet Movie Data Base (IMDb). No formatting is required.\\
\noindent{\textbf{2. Image collection for each identity:}}
Each actor name is then queried in Bing Image Search. Search results
are improved by appending the keywords `actor' and `Sherlock TV series'
to the name of each actor (this avoids erroneous actors with
the same name). Exact duplicate images are automatically removed. Some amount of data is also
obtained automatically from animated GIFs created by fans, providing up to 10
images per GIF. \\
\noindent{\textbf{3. Final manual filtering:}}
This is for two reasons. First, there are erroneous images
(corresponding to the wrong identity), these can trivially be removed (and in fact this could be
automated). However, the second reason is that in order for the power
of the method to be scientifically assessed, images of actors online
which have been taken from the TV show, e.g.\ images of Benedict Cumberbatch
where he is playing the character `Sherlock' must be removed -- this is
purely to create a complete separation of domains (actor -- character).
In practice, this is only required for the more
 obscure actors, and  is a quick and easy process with not more than 20
 minutes of active human effort required for each episode. \\
\noindent{\textbf{4. Augmentation:}}
 Unlike disparities such as facial pose, occlusions and hairstyle, differences between the two domains such as illumination and resolution can be readily resolved using standard data augmentation techniques. Since images available online are usually brighter and of higher quality, the dataset is augmented in three different ways. Every image has a low contrast version, created using contrast stretching (normalization) with limits 0.4 and 1.0, a downsampled version obtained using bicubic interpolation (in an attempt to mimic lower resolution), and a horizontally flipped version. This makes the dataset 4 times larger. 
\vspace{-5pt}
\subsection{Face Context for Character Face Classifier} \label{context}
\vspace{-5pt}
Due to the domain adaptation issues discussed above, we aim to learn
the face context of the character, but not the actor. To achieve this
goal we distinguish between capturing only {\em internal} features
(eyes, nose and mouth) and the extension to {\em external} features
(contour, forehead, hair) by using facial bounding boxes of different
extents.  In the face classification literature, particularly for
surveillance, internal features are found to be more reliable, since
external information is more variable~\cite{masip2007measuring};
similarly, faces in the target domain (TV material) have different
hairstyles and poses from those in the source domain (web images).
Thus, when moving from source to target domain, it is sensible to
restrict the support region for learning the actor face classifiers to
internal features. Within the target domain, however, these external features prove
essential in making the shift from frontal to profile faces, as well
as providing a more robust model for cases where the face is partially
occluded. The different support regions are shown in the
stages 1 and 2 panels of Figure~\ref{fig:overview}.
\vspace{-5pt}
\subsection{Face Track Classification}
\vspace{-5pt}
A similar process is used for each of the three face classifiers.
Features vectors are obtained for each face track using a CNN (see
implementation details~Section~\ref{sec:implementation}).  A linear SVM
classifier is trained for each character, with the images of all other
characters as negative training examples (one-vs-rest approach).  The
tracks are then ranked according to the maximum SVM score, 
and tracks above a certain rank threshold are
assumed to have correct labels. 
\vspace{-5pt}




\section{Speech Modality} \label{Voice}
\vspace{-5pt}
While many approaches to this task focus on recognizing faces, very
few leverage the use of the audio track to assist the automatic
labelling of characters in TV material. As humans, this is an important cue
for recognizing characters while watching TV; particularly if
different characters have distinctive accents or pitch.
Inspired by the success of
speaker identification methods on TV broadcast 
material~\cite{giraudel2012repere}, we aim to create 
voice models for every character. 
 Besides aiding in character identification, such models can be used in conjuction with appearance
  models for `reanimating' or creating virtual avatars of popular TV
  characters~\cite{charles2016virtual}. 
Unlike news broadcasts, this task is more difficult in TV series for the following 
 reasons: first, segmenting the audio track of a show or film into
 regions containing only a single speaker (speaker diarization) is difficult due to rapid shot
 changes, quick exchanges and crosstalk (common due to the
 conversational nature of speech); and, second, the speech segments
 themselves are degraded with a variety of background effects,
 crosstalk, music and laughter tracks. We tackle both these issues in
 the following two subsections.
\vspace{-8pt}
\paragraph{1.\ Speaker diarization.}
Speaker diarization involves segmenting an audio stream into speaker
homogeneous regions, which we henceforth term `speech
segments'. Speech segments contain speech from a single character,
analogous to the face tracks used above. While~\cite{Tapaswi12} impose
the constraint that the speaker is visible in the given shot,
this assumption can be erroneous in the case of `reaction shots', as
well as voiceovers, dubbing and flashbacks. In order to obtain speech
segments, we first extract audio segments corresponding to a single
face track, and then apply Active Speaker Verification (ASV) to
determine if the character in question is speaking. Since the ASV
process is not precise to the frame level, we only use face tracks
where the ASV verifies that the face speaks for the entire track. The implementation
details are given in Section~\ref{sec:implementation}.
\vspace{-8pt}
\paragraph{2.\ Speaker identification.}
Once the speech segments are obtained, speaker identification is used
to build voice models for every speaking character. While most recent
approaches for speaker identification in
TV~\cite{{}charlet2013improving,poignant2015unsupervised} use Gaussian Mixture Models (GMMs) trained on low dimensional features such as MFCCs~\cite{reynolds2000speaker,reynolds1995robust},
we use here CNNs trained on raw spectrograms.

Since the number of speech segments extracted per episode is small
(see Table \ref{table:statistics}),  it is not possible to train a CNN
from scratch. Hence we adopt a feature approach similar to that used
for faces. We use the VoxCeleb CNN which is pretrained on a large speaker identification dataset~\cite{Nagrani17} with raw spectrograms as input. 
This ensures that very little preprocessing is required (see
implementation details~Section~\ref{sec:implementation}). Using this
network, we extract CNN based feature vectors and train a SVM
classifier similar to the classifiers used above. \\ An important
point to note here is that our method requires only speaker
identification (and not speech recognition), and hence works for
videos filmed in any language.
\vspace{-5pt}
\begin{figure*}[ht] 
\centering 
\includegraphics[width=1\textwidth]{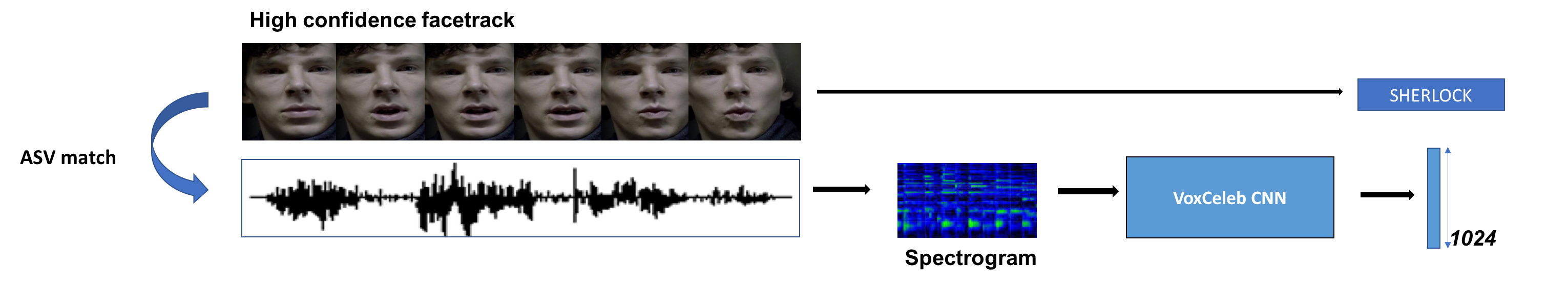} 
\caption{\small{Speaker Identification. The entire speech segment spectrogram is represented by a 1024-D vector, and labels are propagated from the high confidence facetracks.}}
\label{fig:speaker} 
\vspace{-8pt}
\end{figure*}

\section{Implementation Details}
\vspace{-5pt}
\label{sec:implementation}
\subsection{Face}
\vspace{-5pt}
\noindent{\textbf{Face tracking:}}
Face tracking employs the `tracking by detection' approach
used by~\cite{Everingham06a}. This involves five major steps: (i)
frame extraction, (ii) shot boundary detection, (iii) face detection
(iv) face tracking, (v) face/non-face classification. 
While shot boundaries are detected by
comparing colour histograms across consecutive frames, face detection
is done using a faster R-CNN model trained on the large scale WIDER
face dataset~\cite{DBLP:journals/corr/JiangL16a}. Within each detected
shot, face detections are grouped together into face tracks using a
Kanade-Lucas-Tomasi (KLT)  tracker. While most false (non-face) detections are filtered out during this tracking stage, some false positives persist. These
are rejected (at the `track' level) using a face/non face classifier based on
the ResNet-50~\cite{he2015deep} CNN architecture pre-trained on the
VGG Face Dataset~\cite{Parkhi15}, and fine-tuned for the face/non-face
task.  \\
\noindent{\textbf{Face track Descriptors:}}
 Since the faster R-CNN detector used is trained to locate only the internal features, within the target domain the detection bounding boxes are extended by a factor of 0.25 to capture external features as well.  Face track descriptors are then obtained following the approach of~\cite{Parkhi15}. Detected face images are rescaled and passed through the VGG-Face CNN~\cite{Parkhi15}, and the output of the penultimate layer (the last fully connected) is computed to give 4096 dimensional feature descriptors. Feature vectors for every frame in the track are then sum-pooled and L2 normalized, giving a final 4096 dimensional vector for each track. \\
\vspace*{-0.5cm}
\subsection{Voice}
\vspace{-5pt}
\noindent{\textbf{Active Speaker Verification (ASV):}}
Face tracks with less than 50 frames (shorter than 2 seconds) are
discarded as non speaking tracks. ASV is then performed using the
two-stream CNN architecture described in~\cite{Chung16a}, to estimate the correlation between the audio
track and the mouth motion; albeit trained with a slight modification.
Instead of using just cropped images of the lips, the entire face is
used. This allows ASV even for extreme poses. A 1D median filter is
then applied to obtain a confidence score for the track and to
classify the corresponding audio segment as a speech or non-speech
segment. \\
\noindent{\textbf{Audio preprocessing:}} 
Each speech segment is converted into a spectrogram using a hamming
window of width 25ms, step 10ms and 1024-point FFT. This gives
spectrograms of size \(512 \times N\), where \(N = 100L\) , and \(L\) is the
length of the speech segment in seconds.  \\
\noindent{\textbf{CNN based descriptors:}} 
Spectrograms are then passed through the VoxCeleb CNN~\cite{Nagrani17}. This is based on the VGG-M~\cite{Chatfield14} architecture,  modified to have average pooling before the last
two fully connected layers, as well as smaller filter sizes to adapt
to the spectrogram input. Features are extracted from the last fully
connected layer (fc8), and since the entire speech segment can be
evaluated at once by changing the size of the pool6 layer, a single
feature vector of size 1024 is obtained for every speech segment (illustrated in Figure~\ref{fig:speaker} ).

\vspace{-5pt}
\subsection{Face track Classification}
\vspace*{-0.1cm}
After a model (face or voice classifier) is evaluated on the
face tracks, the tracks are then ranked according to the maximum SVM
score, \(\max\limits_{y}w_{y}^{T}x_i+b_y\), where $y$ runs over the
actor/character classes, and tracks above a certain rank threshold are
assumed to have correct labels.  These are then added to the training
set of the next model. \\
\textbf{Face:} The rank threshold for the face classifiers is set manually for episode 1 at 50\% of ranked face tracks, and then applied to all other episodes. \\
\textbf{Speaker identification:} All speech segments are split into train, test and val sets as follows: The train and val segments are those corresponding to confident face tracks with labels propagated from the face stage, and test segments are those tracks with low face confidence. The speaker classifier is extremely accurate, with almost 85\% of test segments identified correctly (ground truths for this evaluation are obtained manually, however are not used in the pipeline). The face tracks corresponding to the top 80\% of test speech segments are then `corrected' with the speaker results.

\vspace*{-0.4cm}
\section{Datasets} \label{Datasets}
\vspace{-5pt}
This section describes the two datasets used to evaluate the method.   \\
\textbf{Sherlock:} 
All 3 episodes of the first season of the TV
series ``Sherlock''. Each episode is about 80 minutes in length. Being
a fast paced crime drama, this is a challenging dataset because it contains (i) a variety of indoor/outdoor
scenes, lighting variations and rapid shot changes; (ii) variations in camera angle and position leading to unusual
viewpoints, occlusion and extreme facial poses; and  (iii)
a large and unique cast as
opposed to popular datasets for this task such as `Big Bang Theory' (which are limited to 6 principal characters).  \\
\textbf{Casablanca:}
Casablanca is a full-length feature film used by Bojanowski {\it et
al.}~\cite{Bojanowski13} for joint face and action labelling. 
Unlike `Sherlock', the video is black and white. We use
face tracks and annotations provided on the
author's website~\cite{Bojanowski13} for evaluation.

Results on both datasets are given in Section~\ref{Results}, and 
statistics are summarized in Table~\ref{table:statistics}. The average number of web images for each actor is less than 100, with more images available for the more famous actors (Benedict Cumberbatch: 147, Ingrid Bergman: 143). Furthermore, there is a gender dependent disparity, with more photos of female actors in circulation, a likely result of a higher propensity for photoshoots/modelling. For Casablanca, the images of the actors downloaded include both greyscale and colored images.

\begin{table}[h!]
\centering
\footnotesize
\begin{tabular}{ l | r | r | r | r  }
  \hline
  \textbf{Episode}   & SE1   & SE2   & SE3   & Casablanca \\ \hline 
  \# Actor images    & 147 / 91.9 / 14 & 147 / 94.3 / 8 & 147 / 85.2 / 13&143 / 41.8 / 5    \\
  \# Face tracks      & 1,731 & 1,775 & 1,740 & 1,273  \\ 
  \# Speech segments & 348   & 345   & 274   & 177    \\ 
  \# Characters      & 15    & 12    & 16    & 17 \\ \hline 
\end{tabular} 
\vspace*{-0.1cm}
\vspace{-5pt}
\normalsize
\caption{\small{Statistics for episodes 1--3 of Sherlock, and for the film Casablanca. Where there are three entries in a field,
numbers refer to the maximum / average / minimum.}}
\label{table:statistics}
\end{table}
\vspace{-5pt}
\vspace{-5pt}
\vspace*{-0.4cm}
\section{Experiments}
 \label{Results}
\vspace*{-0.2cm}
We evaluate the approach on both datasets, and measure performance using face track classification accuracy and Average Precision (AP).
\begin{figure}
\includegraphics[width=0.95\textwidth]{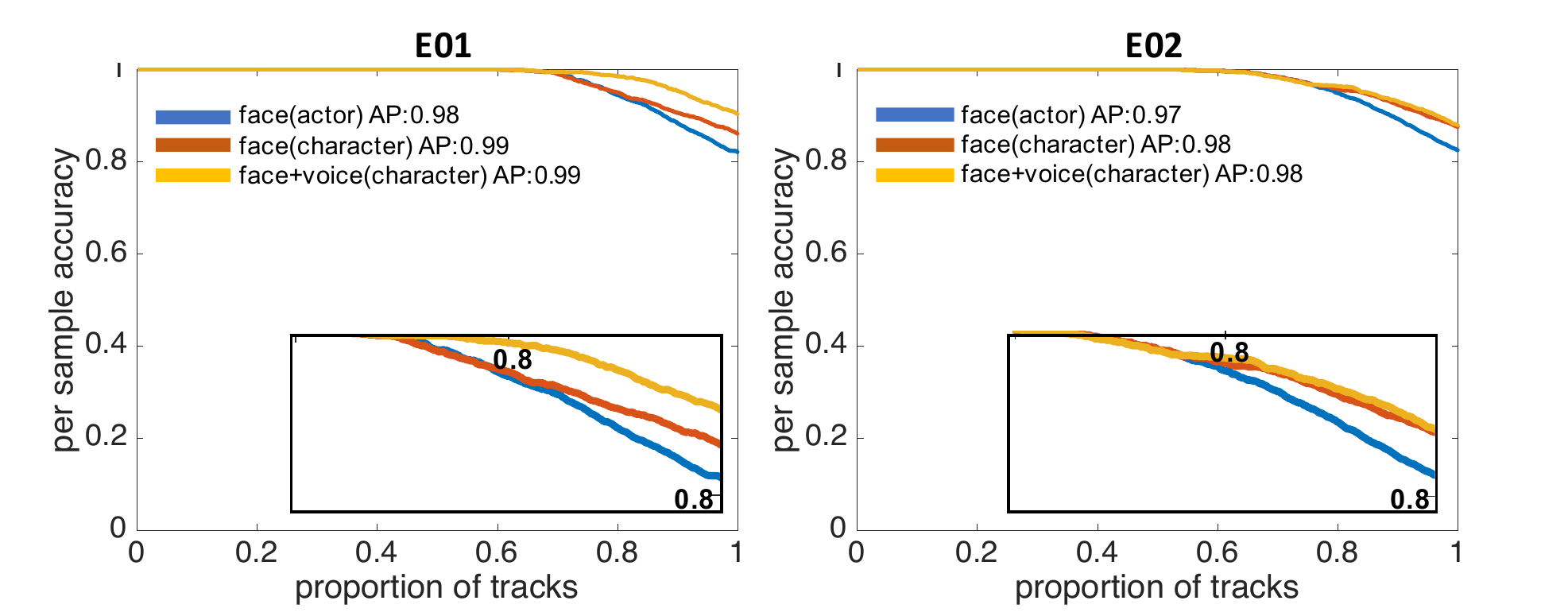}
\vspace*{-0.2cm}
\caption{\small{Precision-Recall Curves for the first 2 episodes of Sherlock. The inset shows a zoomed version of the top right corner.}}
\setlength{\belowcaptionskip}{-50pt}
\label{fig:PR1}
 \end{figure}
 \begin{figure}
\includegraphics[width=0.9\textwidth]{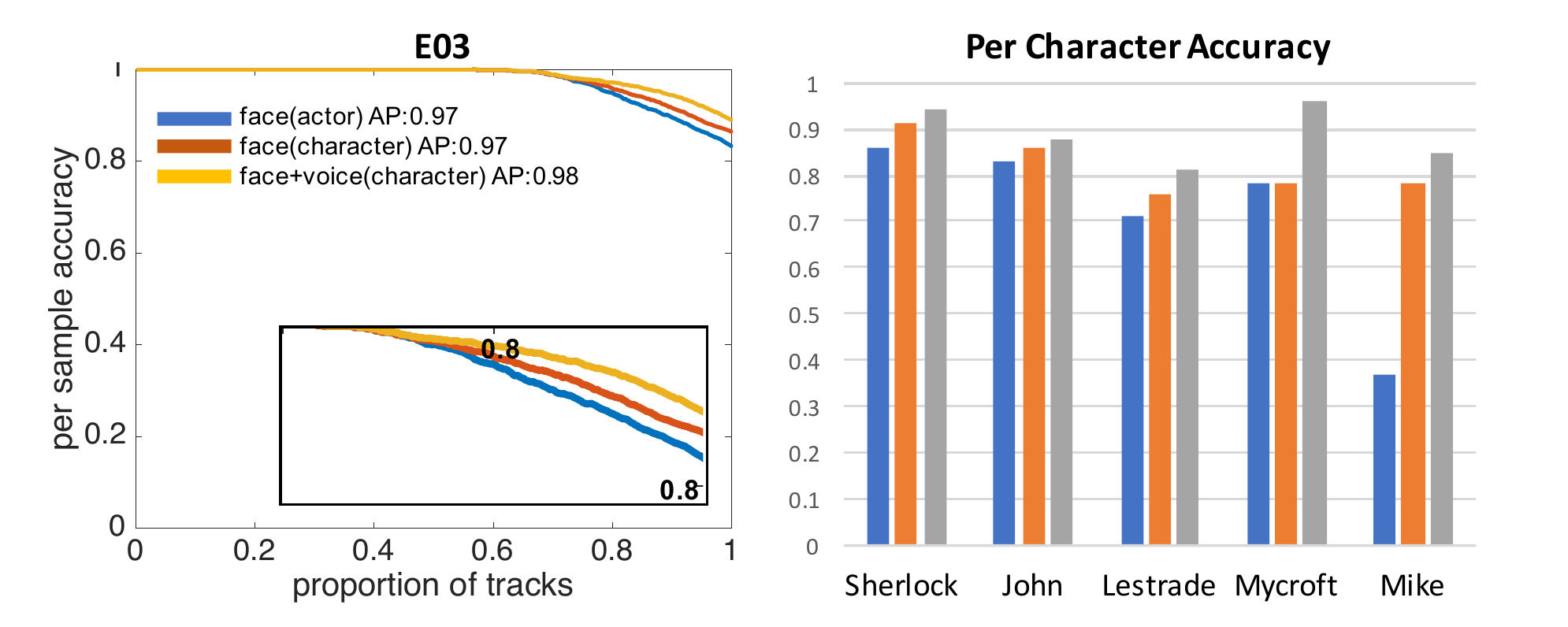}
\vspace*{-0.2cm}
\caption{\small{Precision-Recall Curve for episode 3 of Sherlock (left) and the corresponding character breakdown for this episode (right). Note the large performance increase with the character and voice models for Mike, who has little actor level supervision.}}
\setlength{\belowcaptionskip}{-50pt}
\label{fig:PR2}
 \end{figure}
 \vspace{-5pt}
\begin{table}[!htbp]
\centering
\footnotesize
\begin{tabular}{ l | r | r | r | r |r |r }
\hline
  \textbf{Model}             & \multicolumn{2}{c|}{E1}    & \multicolumn{2}{c|}{E2}   & \multicolumn{2}{c}{E3} \\ \hline 
  {} Face (actor)             & 0.85 &  0.98  & 0.82 & 0.97 & 0.82 & 0.97\\
  \ Face (character context) & 0.88  &  0.99 & 0.86 & 0.98 & 0.85 & 0.97  \\
  \ Face + Speech (character context) &  \textbf{0.92} & \textbf{0.99} & \textbf{0.90} & \textbf{0.98} &\textbf{0.88} & \textbf{0.98} \\ \hline
\end{tabular} 
\vspace{-5pt}
\normalsize
\caption{\small{Classification accuracy (left) and average precision (right) for episodes 1-3 of Sherlock. Accuracy is the more sensitive measure.}}
\label{table:acc}
\end{table}{}

\vspace*{-0.2cm}
\subsection{Results on Sherlock}
\begin{figure*}[ht]
\centering 
\includegraphics[width=0.9\textwidth]{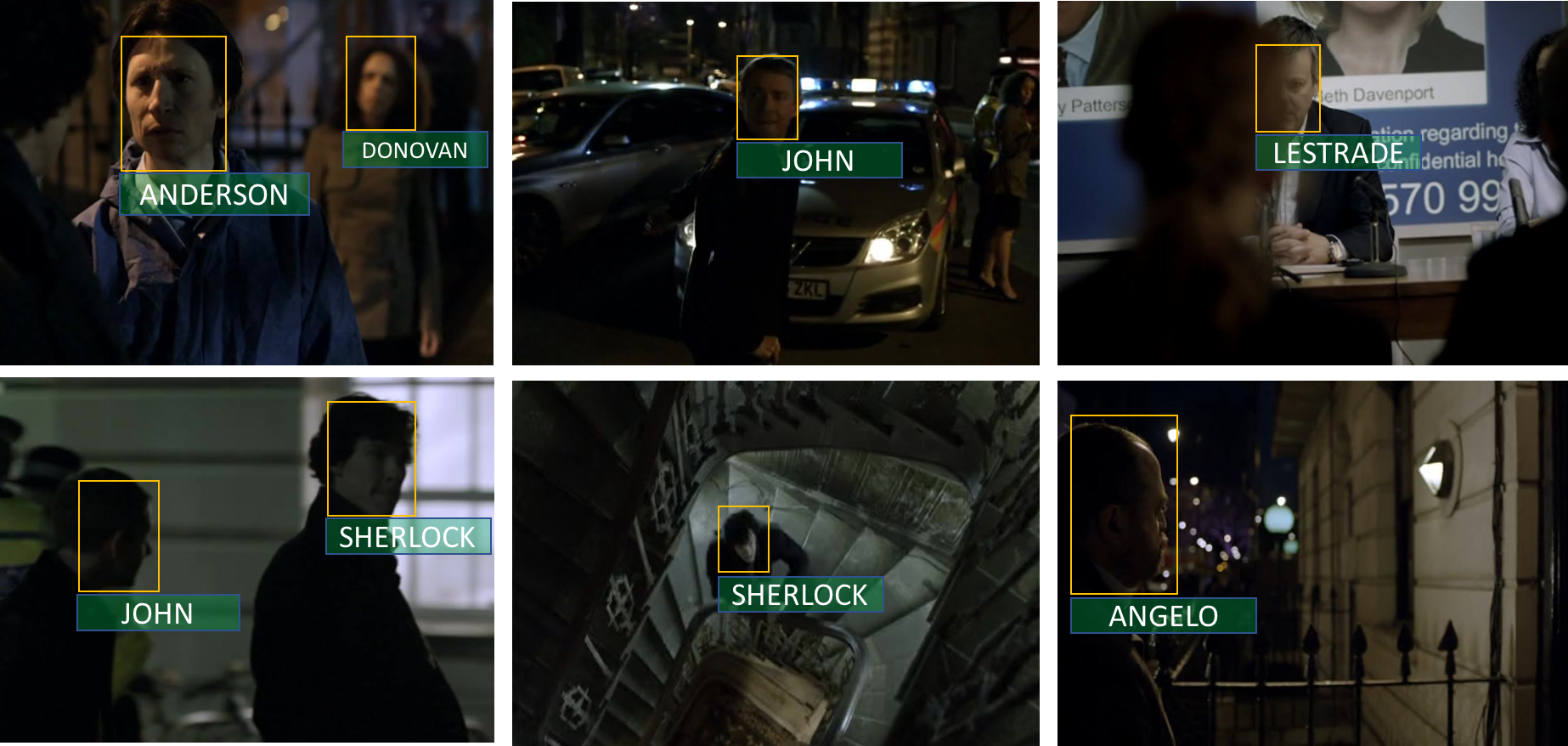} 

\caption{\small{Examples of correct labelling results. From left to right: a blurred face in the background (Donovan), a dark frontal face and a partially occluded face (top row); profiles, extreme pose, and a lesser known actor (bottom row). Note: the original frames have been cropped for optimum viewing.}}

\label{fig:picresults} 
\vspace{-5pt}
\end{figure*}

\vspace{-5pt}
Results for all three face  models are given in Table \ref{table:acc}
and Figures~\ref{fig:PR1} and~\ref{fig:PR2}, with 
example labelling shown in Figure \ref{fig:picresults}.
Overall, the performance is very high, with average precision for episode 1
almost saturated at 0.993.\\
\textbf{Effect of character face models:} 
As can be seen from Table \ref{table:acc}, using the character face
model can provide up to a 5\% classification accuracy boost. This is
particularly important for those actors
where the hairstyles of their character differ greatly from those of
the actor (see Figure~\ref{fig:domains}). \\
\textbf{Effect of voice models:} Voice identification also works very well. It increases performance for two main reasons: 
(1) it helps with the recognition of characters that have large
speaking roles played by lesser known actors, e.g.\ the characters Mike and Mycroft 
(See Figure~\ref{fig:PR2}, right); 
(2) it helps with the identification of
profile tracks of \textit{all} characters for which actor-level
profile supervision is severely lacking.  This can clearly be seen in
Figure~\ref{fig:picresults} (row 2), which shows (from left to right)
examples of a profile face, extreme pose face, and face with
with little actor level supervision. \\ Error cases include heavy occlusions and small dark faces. As seen
from Figures~\ref{fig:PR1} and~\ref{fig:PR2}, all the errors occur at the high recall end
of the curve where the classifier scores are low. This makes it
possible to apply a `refuse to predict' threshold, to avoid
misclassification.

\vspace{-5pt}
\begin{figure}
\begin{floatrow} \CenterFloatBoxes

\ffigbox{%
  \includegraphics[width=6.5cm, height=4cm]{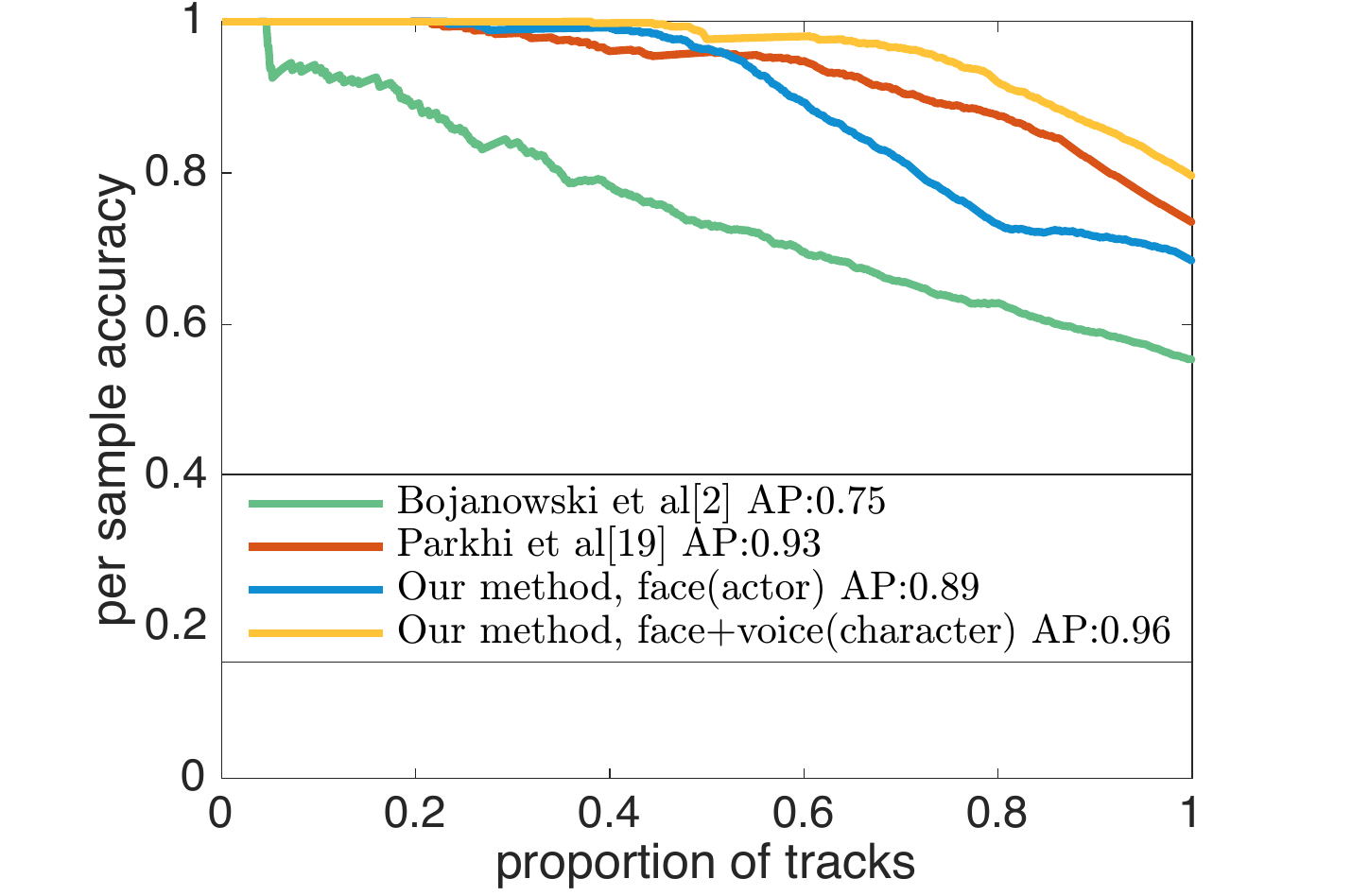}%
\vspace*{-0.1cm}
}{%
  \caption{\small{PR curve for Casablanca.}}%
}\label{figure:PRCas}

\capbtabbox{%
 \footnotesize 
\begin{tabular}{ l | r }
  \hline
  \textbf{Model}                                                      & AP    \\ \hline 
  \ Sivic et al~\cite{Sivic09}, as reported by ~\cite{Bojanowski13}   & 0.63  \\ 
  \ Cour et al~\cite{Cour11}, as reported by ~\cite{Bojanowski13}     & 0.63  \\ 
  \ Bojanowski et al~\cite{Bojanowski13}                              & 0.75  \\
  \ Parkhi et al~\cite{Parkhi15a}                                     & 0.93  \\
  \ Our method                                                        & \textbf{0.96}  \\ \hline
\end{tabular}

\vspace*{-0.1cm} 
}{%
  \caption{\small{Comparison to state-of-the-art.}}%
  \label{table:stateofart}
}

\end{floatrow}
\end{figure}




\vspace{-0.7cm}
\subsection{Comparison with the state-of-the-art} 
\vspace{-5pt}
As noted by~\cite{Bojanowski13}, feature films provide different challenges to
TV series, and hence we modify our approach on the `Casablanca' dataset as follows: \\
\textbf{Actor level supervision:} 
Images of actors available online are both greyscale and in
colour. Hence in the data augmentation step, we convert all images to
greyscale, to match face tracks from the film. 
\\
\textbf{Background characters:} 
Films usually contain both principal characters (who appear often and have speaking
parts), and background characters, who may make a single appearance
and do not speak. While Sherlock has very few background
characters, this is not the case for a feature film like
Casablanca. In the many crowded bar scenes or outside on the streets, there are multitudes of background actors, with more than 30\% of
the tracks belong to the background category.  To
deal with this problem, we adopt the background classifier
of~\cite{Parkhi15a}, a simple SVM based classifier trained
on track level statistics. Using this
classifier, around 20\% of the
background tracks are identified. Tracks classified as background are 
then excluded in stage 1. \\
\textbf{Results:} 
As is clear from Table \ref{table:stateofart}, we beat all previous
attempts on this dataset, even though they use the
stronger supervision provided by a combination of scripts and
subtitles, and~\cite{Parkhi15a} also used VGG-Face CNN features. 
The performance boosts are extremely large
for the principal speaking characters due to voice modelling. 
Most mistakes are for the background characters -- these 
are very difficult to identify, largely because they are not
credited and do not speak. Adding the ability to successfully model
background characters will be an extension of this work.

\vspace*{-0.3cm}
\section{Summary and Extensions}
\vspace{-5pt}
We propose and implement a simple yet elegant approach for automatic character identification in video, achieving high performance on a new dataset, `Sherlock', and exceeding all previous results on the dataset `Casablanca'.
While face and voice models have been successfully created for each character, extensions to this work include training face and speech descriptors in a joint framework to create combined features. The speech modality can also be enhanced by applying ASV precise to the frame (and not track) level, in order to obtain more speech segments. 

By eliminating the need for subtitles, transcripts and manual annotation, our approach is truly scalable to larger video datasets in different languages. We hope that the strength of our results will encourage the community to move towards more transcript-free approaches in the future.  

\paragraph{Acknowledgements.} We are grateful to Qiong Cao and Joon Son Chung for their help with this research.
Funding was provided by the EPSRC Programme Grant Seebibyte EP/M013774/1.

\clearpage
\bibliographystyle{plain}

\end{document}